\documentclass{article}

\usepackage{PRIMEarxiv}
\usepackage{multirow}
\usepackage[utf8]{inputenc} 
\usepackage[T1]{fontenc}    
\usepackage{hyperref}       
\usepackage{url}            
\usepackage{booktabs}       
\usepackage{amsfonts}       
\usepackage{nicefrac}       
\usepackage{microtype}      
\usepackage{lipsum}
\usepackage{fancyhdr}       
\usepackage{graphicx}       
\usepackage{epstopdf}
\graphicspath{{media/}}     

\usepackage[inkscapelatex=false]{svg}
\title{Enhancing Traffic Prediction with Textual Data Using Large Language Models
}

\author{
  Xiannan, Huang \\
  Tongji University \\
  Shanghai, China\\
  \texttt{huang\_xn@tongji.edu.cn}\\
}

\begin{document}
\maketitle

\begin{abstract}
Traffic prediction is pivotal for rational transportation supply scheduling and allocation. Existing researches into short-term traffic prediction, however, face challenges in adequately addressing exceptional circumstances and integrating non-numerical contextual information like weather into models. While, Large language models offer a promising solution due to their inherent world knowledge. However, directly using them for traffic prediction presents drawbacks such as high cost, lack of determinism, and limited mathematical capability. To mitigate these issues, this study proposes a novel approach. Instead of directly employing large models for prediction, it utilizes them to process textual information and obtain embeddings. These embeddings are then combined with historical traffic data and inputted into traditional spatiotemporal forecasting models. The study investigates two types of special scenarios: regional-level and node-level. For regional-level scenarios, textual information is represented as a node connected to the entire network. For node-level scenarios, embeddings from the large model represent additional nodes connected only to corresponding nodes. This approach shows a significant improvement in prediction accuracy according to our experiment of New York Bike dataset.
\end{abstract}


\section{Introduction}
Traffic prediction refers to the anticipation of future usage levels of transportation within various zones of a city over a specified period, such as one hour to several hours ,for example, forecasting the usage of shared bicycles in different girds within the forthcoming one hour. This predictive task is crucial for the rational scheduling and allocation of transportation supply. By engaging in more refined and judicious scheduling, it is possible to maximize the operational returns for transportation providers and concurrently enhance the service quality for users.

The research into short-term traffic prediction typically follows a paradigm where historical data, along with contextual information such as weather and time, are inputted into a model to generate future predictions, and the contextual information are always embedded as vectors. This paradigm adheres to the framework of numerical inputs and outputs. However, this modeling approach exhibits deficiencies. Firstly, it lacks the capacity to address exceptional circumstances, such as extreme weather events or special events like concerts or sports matches in certain areas. These exceptional circumstances are inadequately represented in the training data, making it challenging for models to capture their patterns and generalize to testing data. Secondly, many pieces of information are not inherently numerical, such as weather data. While binary embeddings for weather conditions like sunny or rainy are commonly utilized in most models, finer-grained weather information such as air quality , specific temperature values and precipitation typically exist in textual formats. Integrating such textual contextual information into models as embeddings poses challenges. Therefore, handling background information in textual form presents difficulties for conventional numerical-input-output models. Indeed, these two challenges can be closely intertwined, as most exceptional circumstances are also provided through textual information, for instance, holiday schedules and the occurrence of events like concerts.

These two issues can indeed be addressed using large language models. Firstly, the training corpus of large models encompasses many textual data worldwide \cite{10.5555/3495724.3495883}. This implies that these large models have inherent knowledge about certain exceptional circumstances, such as concerts or extreme weather conditions. Even if the training data does not include instances of specific events, these models still retain the capability to generalize to such scenarios. For example, when prompted about the impact of heavy rainfall on taxi demand, ChatGPT can provide insights indicating that heavy rainfall may cause some individuals who initially intended to use public transit or subway systems to opt for taxis instead. This demonstrates that ChatGPT has a basic understanding of the effects of heavy rainfall on taxi usage, as it has been exposed to relevant information during training. Thus, if we employ ChatGPT to forecast taxi demand during heavy rainfall, it can generalize its knowledge to predict demand under such conditions, even if the training data lacks explicit examples of heavy rainfall. Similarly, handling textual contextual information is a bless of large language models, as they are inherently designed to process textual data.

Recently, there has been a surge in efforts to utilize large language models for modeling traffic volume or, more broadly, as a means of modeling time series data. Some studies used traffic volume data to fine-tune pre-trained large language models, and obtained models capable of predicting traffic volume \cite{Liu2024SpatialTemporalLL}. However, such an approach first requires fine-tuning a large language model, which is highly costly. Additionally, the fine-tuned model requires input data in numeric format, thus unable to incorporate background information existing in text format. There are other work treating the traffic volume of individual regions as a time series, feeding its historical data into large models along with some contextual information, and using these models to predict future traffic volume \cite{Xue2022PromptCastAN,Gruver2023LargeLM}. Due to large models not being specifically tailored for traffic prediction, they may not adapt well to such tasks, resulting in lower accuracy. Additionally, the uncertainty of large model outputs means that different queries or even the same prompt may yield varying results across repeated experiments. This lack of determinism is disadvantageous for achieving a certain level of certainty in traffic prediction. Moreover, a more crucial issue is that large model outputs cannot be specified in a particular format. For instance, requesting a direct prediction from the model may not yield a straightforward result; instead, it may provide an analysis or even state its inability to make a prediction, thereby constraining its future utility. Furthermore, a more significant problem lies in the relatively limited mathematical capability of large models and their lack of sensitivity to numerical data. Many studies have highlighted this issue, attributing it to the misalignment between the representation of text and numerical data within large models \cite{Gruver2023LargeLM}. In other words, large models treat numerical data as strings, which compromises their mathematical proficiency.

Hence, many further studies emphasize the utilization of specialized modules to align the textual and numerical spaces \cite{Li2024UrbanGPTSL}. For instance, recent work proposes employing an existing spatiotemporal forecasting model, such as STGCN, as a feature extractor to capture spatiotemporal information. Historical data is passed through this feature extractor to obtain a feature vector, which is then concatenated with background information in the prompt to form the final input prompt for the large model, enabling it to produce output results. However, a challenge here is that such approaches may also require fine-tuning of the large model, which incurs high costs due to the considerable number of parameters in large models and necessitates a substantial amount of data. Additionally, this method is susceptible to the aforementioned drawbacks, such as the inability of different large models to output results in the desired format and the lack of consistency in their outputs.

In this study, we propose a novel and concise approach. Specifically, we utilize large language models to process textual information but refrain from directly using the large model for prediction. Instead, we obtain a textual embedding from the large model for the textual information. This encoding is then combined with historical traffic data and fed into a traditional spatiotemporal forecasting model to obtain prediction values. 

In this study, we investigate two types of special scenarios: regional-level and node-level special scenarios. The former refers to situations that impact the entire city, such as holidays and extreme weather, while the latter pertains to specific locations within the city, such as sporting events or concerts. 

Our approach is straightforward. For regional-level special scenarios, we obtain a vector representation of the background information through text and treat it as a node. This node is connected to all other nodes in the entire network, representing the influence of the special scenario on all regions. This vector representation is then inputted into the large model. As for node-level special scenarios, we obtain embeddings of the textual information through the large model and treat them as additional nodes. However, these additional nodes are only connected to their corresponding nodes, adding only one edge. Through these two approaches, we observe a significant improvement in the final prediction accuracy.
\section{Method}
\subsection{Problem definition and notation}
Supposed there are $n$ districts in a city, and in each time interval, there are $d$ diamentional (such as bicycle inflow and outflow) features in each district. The feature of each districts in the past $t_1$ time intervals are known, donated as $X \in R^{n \times d \times t_1}$.
Besides some contextual information is avilible, and these information can be catagalized as two classes, one is the city-level information such as holiday, weather, air quality and so on, the other is district-level information such as there will be concert or sports game in a certain district. These two information are donated as $i_1,i_2$, respectiely. In general, these information are given in a text format. Ferthermore, the spatial relationship of these districts are given by an adjacent matrix $A \in R^(n \times n)$. The task is to predict the features of each districts in the future $t_2$ time interval, donated as $Y \in R^{n \times d \times t_2}$. In other words, the task in to find a function $f$ to minimize the prediction error of $Y$ as \ref{eq1} shows
\begin{equation}
   f=argmin_f  Loss(Y,f(X,A,i_1,i_2)
   \label{eq1}
\end{equation}
where $f$ is $(X,A,i_1,i_2) \mapsto  Y$.

Besides, the purpose of this work is to refine a traditional traffic prediction model to consider textual information, so we assume the traditional traffic prediction model as $f_t$:
\begin{equation}
    f_t: (X,A) \mapsto  Y
\end{equation}

\subsection{Intergrating regional-level information}
In our article, regional-level background information is primarily considered, encompassing holidays and weather, such as air quality, clear or rainy weather, precipitation levels, and temperature. These details are presented in textual format. For instance, "Today is May 13th, with zero precipitation, zero air quality index, and a temperature ranging from 21 to 30 degrees Celsius. Moreover, today is Sunday." Such textual background information needs to be incorporated into our model for prediction. This task poses a considerable challenge since vector-form inputs, such as historical traffic volume, are utilized by our model, but the background information is textual in nature. To address this, large models are employed to transform text into vectors, known as text embeddings. Typically, the text is inputted into the large language models, and the output vector from the last layer is extracted, utilizing it as the representation or embedding of the text. This technology finds extensive applications in natural language processing tasks, such as text classification and information extraction, as the large models predict the next token based on text representation, allowing the output of the last layer to effectively summarize the input text while incorporating semantic information from the input texts.

Usually, the output embedding vectors are of high dimensionality, often reaching several hundred or even thousands of dimensions. However, since our inputs pertain to specific aspects such as weather or temporal holidays, many dimensions are unnecessary. Thus, dimensionality reduction, specifically principal component analysis (PCA), is employed. Following PCA, the reduced dimensionality is determined based on the principle of retaining 95\% of the variance.

It is emphasized that our approach is plug-and-play, meaning it can be seamlessly integrated into various models. To achieve this, an additional node, termed an auxiliary node, is introduced, which is connected to all other nodes. The feature of this node is the vector of LLM embeddings after PCA. 

We assume the output vector of LLM embedding after dimension reduction is $c \in R^{d_c}$. According to the Problem definition part, the input history record is $X \in R^{n \times d \times t_1}$, and the record of each is of shape $R^{d \times t_1}$. It is needed to convert $c$ into a vector of shape $R^{d \times t_1}$ and concatenate it with $X$, so the concatenated vector can be seen as a graph of $n+1$ nodes and be put into traditional traffic prediction model. To accomplish that, we define $t_1$ linear transform layers, the i-th layer is as follows:
\begin{equation}
    c_i=\sigma(W_i x+b_i)
\end{equation}
Where $W_i,b_I$ are the weight and bias of the i-th layer and with the shape $R^{d_c,d_1},R^{d_1}$ respectively. So the shape of each $c_i$ is $R^{d_1}$. Then all $c_i$ can be concatanated as a vector of shape $R^{d \times t_1}$.
\begin{equation}
    node_c=concat([c_1,c_2,...,c_{t_1}])
\end{equation}
And $node_c$ be consider as the record of an additional node. Then we concatenate it with $X$ and get the expended record $X_e$.
\begin{equation}
    X_e=concat([X,node_c])
\end{equation}
After that, the adjacent matrix is expended, specificly, a row and a column with all one are added to the position of $node_c$, and the expended adjacent matrix $A_e$ can be obtained. Finally, the expended record $X_e$ and expended adjacent matrix $A_e$ are feeded into traditional traffic prediction model to get prediction.
\begin{equation}
    y=f_t (X_e,A_e)
\end{equation}
\subsection{Intergrating node-level information}
\begin{figure}[ht]
    \centering
    \includegraphics[width=\textwidth,height=0.5\textwidth]{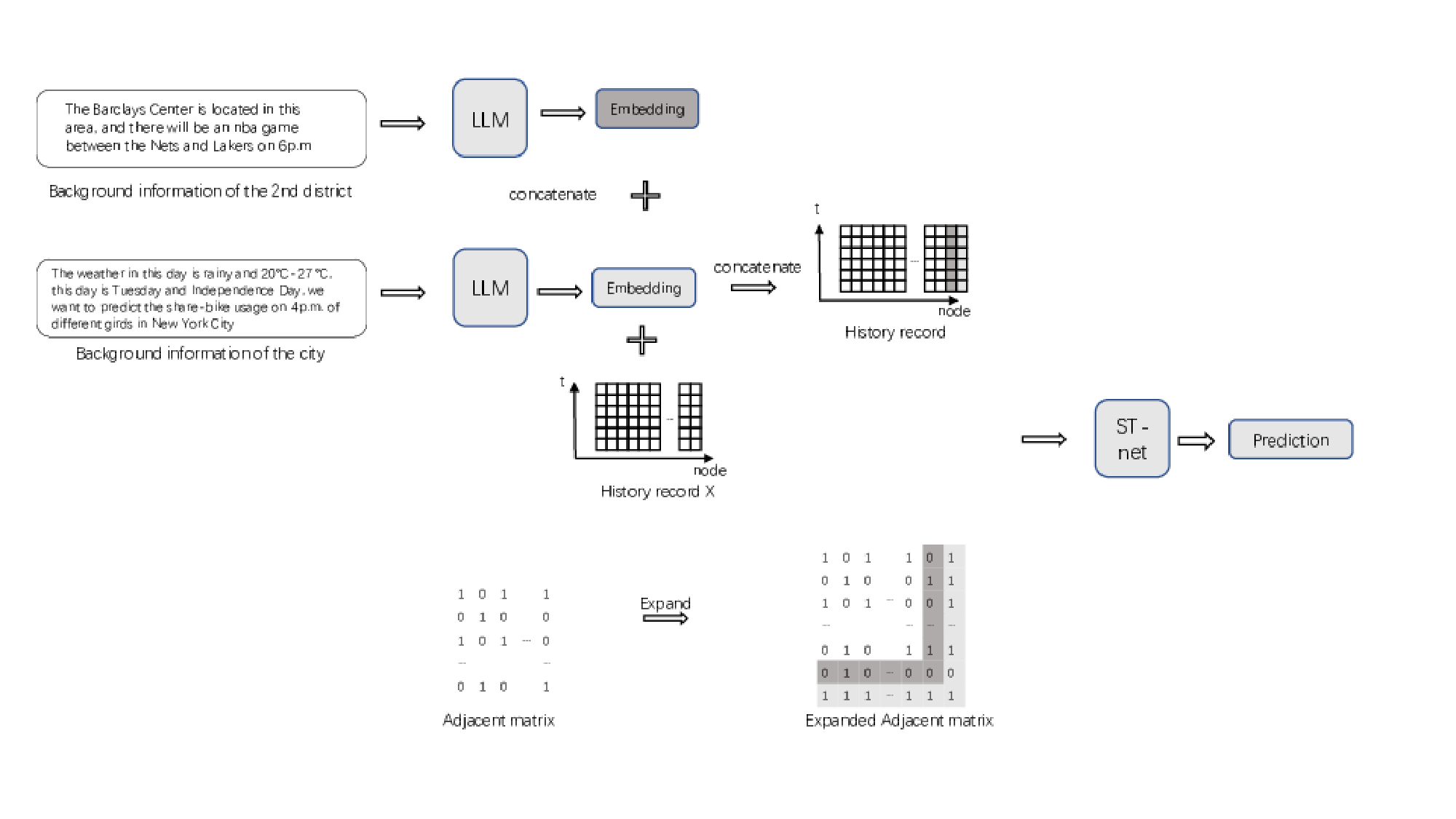}
    \caption{Method Overview.
}
    \label{figq}
\end{figure}
For node-level information, our method closely parallels the approach used for region-level case. Specifically, the background information for the district is arranged as text, which is then processed through a large model to derive the final embedding vector. Similarly, various embedding vectors can be obtained for different types of background information. Subsequently, these embedding vectors undergo dimensionality reduction through principal component analysis, and the resulting reduced vectors are integrated into the model as features.

Similar to the method employed for region-level, the incorporation of features into traditional prediction model involves the addition of an auxiliary node to the graph. And this auxiliary node is specifically linked to the district of interest, indicating its connection to only one node, unlike the city-level case. 

In summary, our method is intuitive and relatively straightforward. By utilizing auxiliary nodes, features are incorporated into the traditional traffic prediction model. The city-level auxiliary nodes are connected to all nodes, while the district-level auxiliary nodes are linked only to the nodes representing the districts of interest. The features of these auxiliary nodes are derived from the embedding vectors obtained from the large model. Subsequently, this data can be inputted into traditional spatiotemporal neural networks to obtain the final predictions. The overall workflow of our method is depicted in Figure \ref{figq}.

\section{Case Study}
\subsection{Date}

\begin{figure}[ht]
    \centering
    \includegraphics[width=4 in]{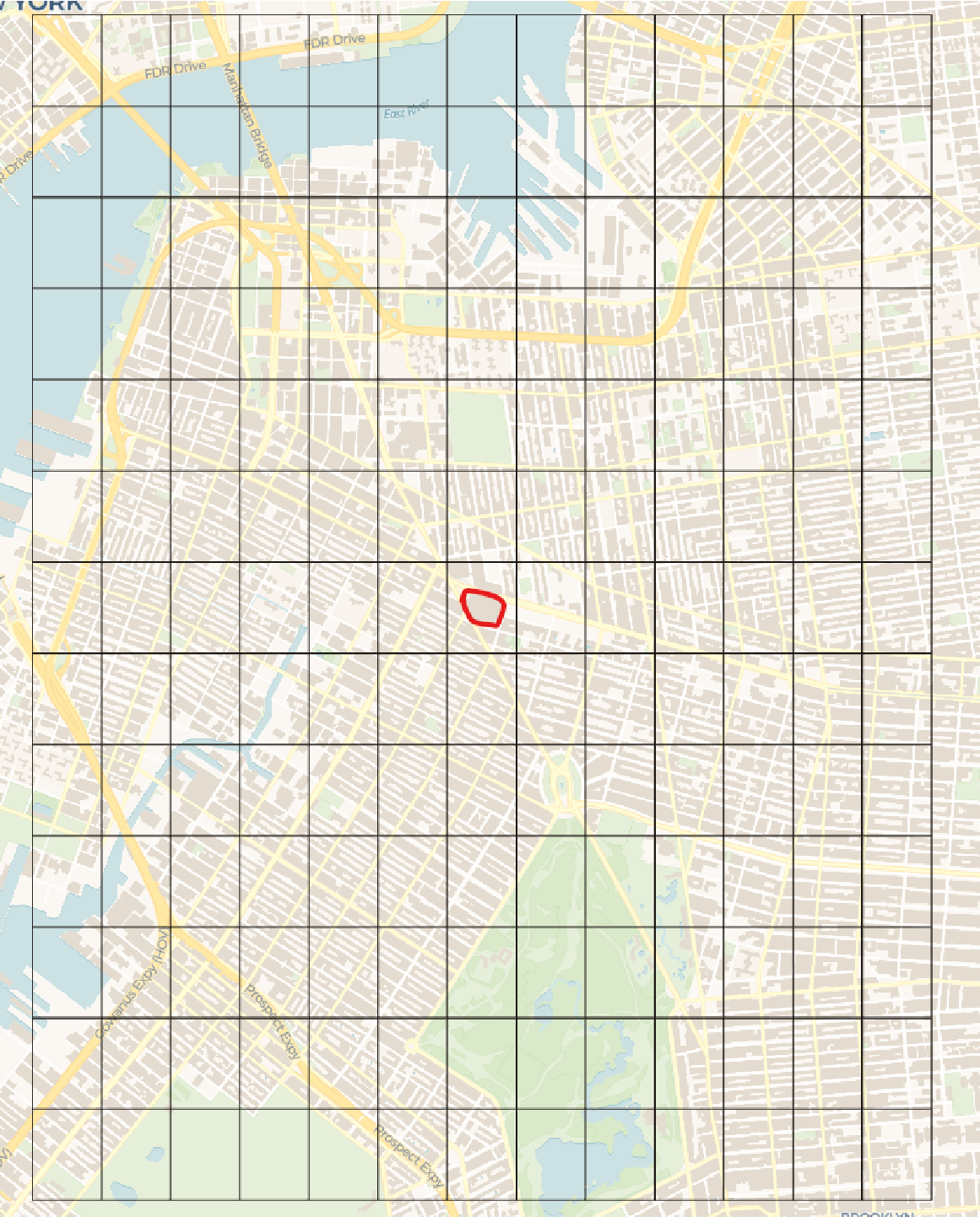}
    \caption{The grid partitioning scheme. The grid marked at the center represents Grid 84, with the Barclays Center in New York highlighted by the red circle.}
    \label{fig:grid}
\end{figure}

\begin{figure}[ht]
    \centering
    \includegraphics[width=\linewidth]{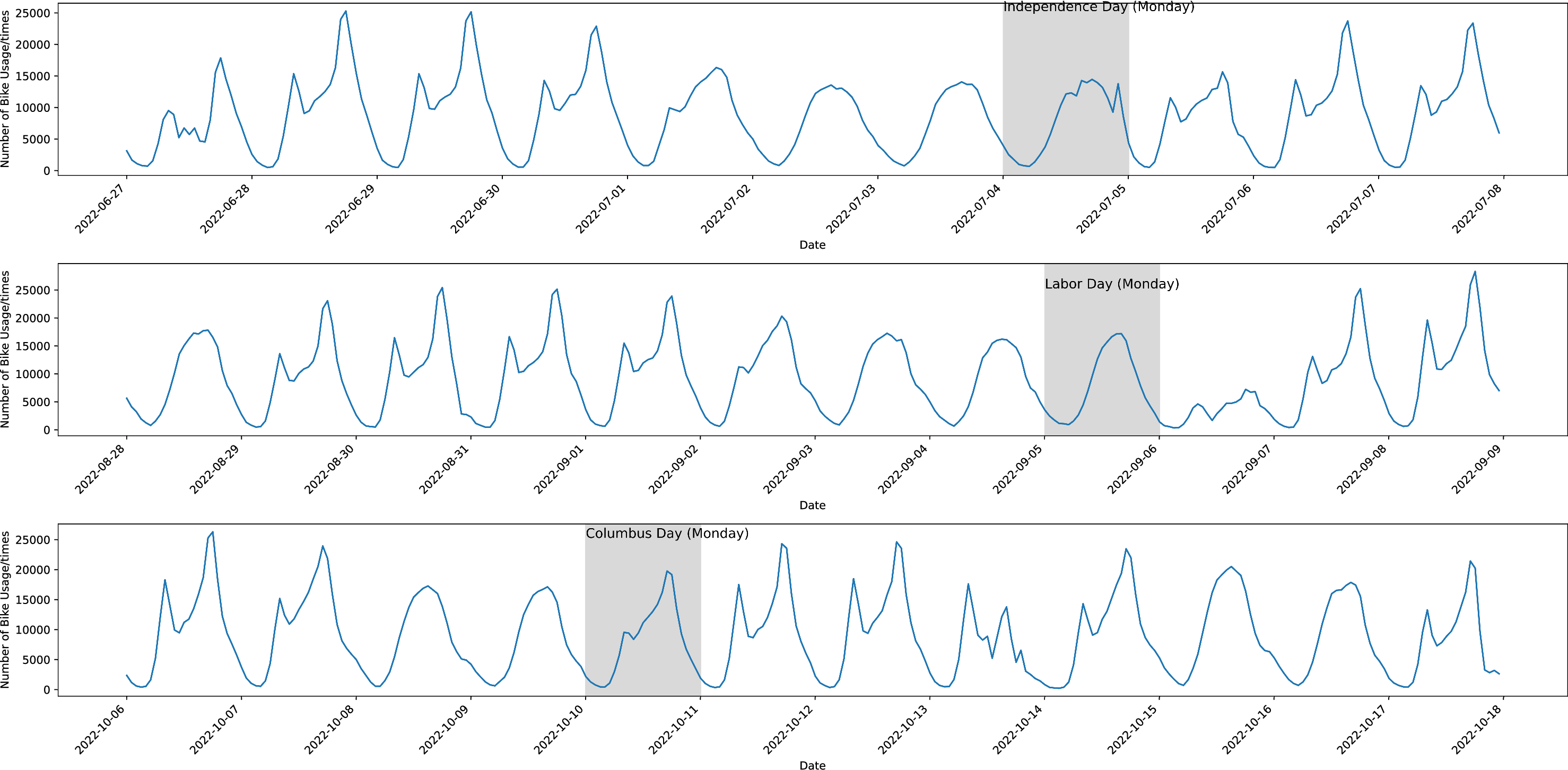}
    \caption{Bike flow in some days}
    \label{fig:enter-label}
\end{figure}

The dataset utilized in our case study consists of New York City's bicycle-sharing data, a widely employed dataset sourced in traffic prediction. Each record within this dataset documents a single trip, including its occurrence time, start point, and end point. The areas of interest are divided into 169 grids, each spanning 1km by 1km, as shown in figure \ref{fig:grid}. Subsequently, based on the usage records of bicycle trips, the bike pickup and drop-off quantities for each grid per hour can be derived. This data forms the basis of our dataset. Specifically, our prediction task is forecasting the bike pickup and drop-off quantities for the respective area for the forthcoming hour, based on the bike pickup and drop-off quantities for each grid over the past six hours, alongside the aforementioned background information, including city-level factors such as air quality, rainfall, temperature, the hour of prediction, the day of the week, and holiday information, 

Additionally, in Figure \ref{fig:enter-label}, the usage of shared bikes in New York City across various hours on specific days is illustrated. Notably, three special days: Independence Day, Labor Day, and Columbus Day, are highlighted. Despite these holidays are all Mondays, they exhibit distinct usage patterns compared to adjacent Mondays. The usage fluctuation graphs for these days can serve as additional evidence supporting the necessity of considering special factors such as holidays.

Moreover, a particular grid, labeled as Grid 84, which is highlighted in our map, as shown in figure \ref{fig:grid}, holds significance due to its proximity to the Barclays Center in New York City, where frequent events such as concerts and sports games, including NBA matches, are held. It is believed that such special events could influence traffic volume. Therefore, information about these events obtained from the Barclays Center's official website is incorporated as additional background information for this specific area.
Consequently, the focus of our experiment lies in the city-level assessment, which encompasses all 169 grids, and the region-level evaluation, which specifically targets grid 84. Our training dataset spans from June 1, 2023, to August 7, 2023, covering a duration of 14 weeks. This is followed by a validation set extending from August 8, 2023, to August 22, 2023, totaling 2 weeks, and finally, a testing set ranging from August 23, 2023, to September 20, 2023, lasting for 4 weeks.

\subsection{Baseline Model}
We selected the following eight fundamental spatiotemporal prediction models to conduct our experiment : AGCRN \cite{AGCRN} DCRNN \cite{li2018diffusion} STGCN \cite{STGCN} MTGNN \cite{MTGNN} GWNET \cite{xu2018graph} TGCN \cite{TGCN} STTN \cite{Xu2020SpatialTemporalTN} STSGCN \cite{Song2020SpatialTemporalSG}. Firstly, we employed their traditional versions for prediction. Subsequently, we utilized the versions enhanced with embeddings from large language model to to predict.

\subsection{Result}
As previously outlined, attention is directed towards two tasks. The first task entails assessing whether prediction accuracy is improved for all grids after the incorporation of enhancements. The second task focuses on whether prediction accuracy is enhanced for a designated grid after the provision of relevant information pertaining to that grid. In our experiments, Grid 84, containing the Barclays Center, serves as the designated grid. Table  \ref{tab:my_label} below presents the prediction errors, MAE and RMSE, for both the original versions of the models and the versions enhanced with embeddings from the large language model. Additionally, errors are reported for all regions and specifically for region 84. It is observed that after the incorporation of embeddings from the large language model, the prediction errors of the models decrease for both tasks.
\begin{table}[!ht]
    \centering
    \begin{tabular}{ccccccccc}
    \toprule
       \multirow{3}{*}{Model} & \multicolumn{4}{c}{Original} & \multicolumn{4}{c}{with LLM} \\
       ~  &\multicolumn{2}{c}{Error in all Regions} &\multicolumn{2}{c}{Error in grid 84} &\multicolumn{2}{c}{Error in all Regions} &\multicolumn{2}{c}{Error in grid 84}  \\ 
       ~ & MAE &RMSE & MAE &RMSE & MAE &RMSE & MAE &RMSE\\ \hline
       AGCRN & 1.911& 2.702& 1.881&2.974&1.739 &2.394 &1.397 &2.523 \\
        DCRNN &2.034& 2.863&1.927& 3.043&1.840 &2.593 &1.430 &2.581 \\
        STGCN&1.913& 2.729& 1.850& 2.953&1.734 &2.470 &1.382 &2.492 \\
        MTGNN&1.924& 2.717&1.820& 2.911&1.692 & 2.523&1.332 &2.473 \\
        GWNET&1.936& 2.711&1.902&3.056&1.759 &2.539 &1.347  &2.538\\
        TGCN&2.296& 3.388&2.288& 3.536&1.932 &2.983 &1.834  &3.193\\
        STTN&1.983& 2.797&1.982&3.139& 1.776&2.675 &1.522  &2.742\\
        STSGCN&2.218& 3.205&2.424& 3.867&1.824 &2.837 &2.043  &3.479\\
     \bottomrule
    \end{tabular}
    \caption{Error for all regions and the specific region of all models with and without LLM embeddings. The 2nd to 5th columns shows the prediction error using original models and the 6th to 9th columns shows the prediction error using original models with LLM embeddings}
    \label{tab:my_label}
\end{table}
\section{Related Work}
Due to the nature of traffic prediction as a spatiotemporal forecasting problem, it involves both temporal and spatial dimensions. Consequently, it encompasses two main analytical aspects: time series analysis and graph-based data analysis. Therefore, in the literature review section, we firstly review articles employing large language models for time series analysis, followed by works utilizing large models for graph-based modeling. Finally, we summarize the relevant literature concerning the application of large models in the field of transportation.

\subsection{LLM for time series analysis}
    Articles predicting time series using large language models can be classified into several categories. The first category involves directly feeding the historical values of the time series into the large language model, enabling the model to output predictions in textual form \cite{Xue2022PromptCastAN,Jin2024TimeSF,Yu2023TemporalDM,Liu2024LSTPromptLL}. The second category involves adopting the architecture of large language models, specifically the Transformer architecture, along with the autoregressive training method, to construct models tailored for time series analysis \cite{Garza2023TimeGPT1,Zhou2023OneFA,Bian2024MultiPatchPA}. In such models, both input and output are vectors rather than text. Additionally, some articles propose embedding traditional models into time series data before utilizing large language models for prediction. This can involve fine-tuning a traditional time series model to align the representations of the time series with those of the large model's space \cite{Liu2024AutoTimesAT,Duan2023DeWaveDE}. Moreover, there is a subset of work exploring the use of multimodal models, time series data is plotted into images, such as line graphs, which are then fed into visual models for prediction \cite{Wimmer2023LeveragingVM,Liu2023ETPLT}.Interested readers may refer to recent reviews for more articles on the application of large models in time series analysis \cite{Zhang2024LargeLM}.As previously mentioned in the introduction, these methods primarily address univariate time series data. Handling high-dimensional data directly with large language models is challenging due to the inability to input relationships between various time series. Given the spatiotemporal nature of traffic prediction task, attention must also be paid to the relationships between different dimensions of time series data across various regions. 
\subsection{LLM for graph}
    The primary challenge in applying large models to graph data lies in their inability to directly comprehend the structural information of graphs. Some methods were proposed to translate graph structural information into text and feed them to LLMs \cite{fatemi2024talk,Zhao2023GraphTextGR}
    for some prediction or classification task. In addition, some methods propose utilizing large models to obtain embeddings for nodes and links, and then traditional graph models, such as GNNs, are applied on the base of these embeddings to ultimately classify or predict graph nodes \cite{liu2024one,Chen2023ExploringTP}. Additionally, some studies propose utilizing some embeddind models (such as GNN) to encode each node in a graph and obtain embeddings for these nodes, which are then fed into large models \cite{Chen2024LLaGALL,Tang2023GraphGPTGI}. Specifically, the input to large models consists of prompts indicating the task, and the features derived from embedding models. In summary, when utilizing large models for tasks in graphs, it is imperative to employ specialized modules to ensure that the models incorporate structural information of the graphs. 
    \subsection{LLM for traffic}
    Since the proposal of large language models represented by ChatGPT and their widespread application across various domains, many scholars in the transportation field have endeavored to apply large language models in transportation-related studies. For instance, researchers have utilized large language models to predict taxi usage during special events \cite{Liang2023ExploringLL}. However, it is important to emphasize that this study differ from ours, as they directly query the model and solely predict a single variable time series, such as the number of taxis in a certain district, without considering spatial relationships.

In the field of autonomous driving, some articles input the current state of the vehicle, its surroundings, and data captured by monitoring cameras into multimodal models, enabling the large model to make decisions about  the next steps in autonomous driving \cite{Ding2023HiLMDTH,Dewangan2023Talk2BEVLB,Tanahashi2023EvaluationOL,Zhou2024GPT4VAT}. Besides, some work proposed using LLM as agent to build autonomous simulation agent \cite{Jin2023SurrealDriverDG} or generate traffic scene for simulation \cite{Tan2023LanguageCT}. Here, the superior logical reasoning capabilities of large models compared to conventional models are utilized, rendering them suitable for autonomous driving tasks. Furthermore, some studies propose fine-tuning large models as intelligent agents for city simulations, adapting them to urban environments. Researchers highlight this approach allows large models to simulate unique intelligent agents with specific characteristics \cite{Xu2023UrbanGI}. Additionally, some studies utilize large language models to analyze textual information related to transportation, which was challenging to model with traditional methods before the advent of large models \cite{Zhang2024TransportationGamesBT}. For example, researchers analyze textual descriptions of traffic accidents using large language models to determine the causes of accidents and identify responsible parties \cite{Mumtarin2023LargeLM}. Other studies analyze users' social media posts using large language models to infer their attitude towards public transportation \cite{Momin2024LeveragingSM}. Moreover, some articles propose using large language models to predict users' next visited locations \cite{Wang2023WhereWI} or to analyze users' potential transportation choices based on their personal attributes and the costs and utilities of various transportation modes \cite{Mo2023LargeLM}. There are also some work tried to use LLMs to manuipulate traffic software and interact with user with nature language \cite{Zhang2023TrafficGPTVP,Da2023OpenTIOT}.

In summary, literature across various aspects of the transportation field has explored the application of large language models. Compared to traditional transportation models, large language models offer considerable flexibility, as they can analyze various types of textual data rather than being limited to numerical inputs. Additionally, large language models exhibit strong generalization capabilities, enabling them to adapt to specific situations, as discussed in the introduction. Moreover, large models possess superior logical reasoning capabilities compared to conventional models, making them suitable for decision-making tasks such as in autonomous driving.
\section{Conclusion}
This study introduces an innovative approach to traffic prediction that significantly enhances the accuracy of forecasting, particularly under atypical conditions such as extreme weather or special events, by integrating textual and numerical data through the use of Large Language Models. The core of our method lies in transforming textual information, such as weather descriptions and holiday information, into embedding vectors that are then combined with historical traffic data for input into traditional spatiotemporal forecasting models.

A case study utilizing New York City's bike-sharing dataset has demonstrated the advantages of our proposed method across various level traffic prediction tasks. Experimental results indicate that the incorporation of LLM embeddings has led to a reduction in predictive errors for both overall regions and specific areas, such as Grid 84 where the Barclays Center is located, affirming the effectiveness of our approach.

In summary, the method presented in this study offers a fresh perspective in the domain of traffic prediction. By combining textual and numerical data and leveraging the advanced capabilities of large language models, we can significantly improve the accuracy and reliability of forecasts. Future work can further investigate how to optimize the representation of embedding vectors and explore the application of this method to a wider range of traffic prediction scenarios and tasks.
\bibliographystyle{unsrt}  
\bibliography{references}

\begin{thebibliography}{10}

\bibitem{10.5555/3495724.3495883}
Tom~B. Brown, Benjamin Mann, Nick Ryder, Melanie Subbiah, Jared Kaplan, Prafulla Dhariwal, Arvind Neelakantan, Pranav Shyam, Girish Sastry, Amanda Askell, Sandhini Agarwal, Ariel Herbert-Voss, Gretchen Krueger, Tom Henighan, Rewon Child, Aditya Ramesh, Daniel~M. Ziegler, Jeffrey Wu, Clemens Winter, Christopher Hesse, Mark Chen, Eric Sigler, Mateusz Litwin, Scott Gray, Benjamin Chess, Jack Clark, Christopher Berner, Sam McCandlish, Alec Radford, Ilya Sutskever, and Dario Amodei.
\newblock Language models are few-shot learners.
\newblock In {\em Proceedings of the 34th International Conference on Neural Information Processing Systems}, NIPS '20, Red Hook, NY, USA, 2020. Curran Associates Inc.

\bibitem{Liu2024SpatialTemporalLL}
Chenxi Liu, Sun Yang, Qianxiong Xu, Zhishuai Li, Cheng Long, Ziyue Li, and Rui Zhao.
\newblock Spatial-temporal large language model for traffic prediction.
\newblock {\em ArXiv}, abs/2401.10134, 2024.

\bibitem{Xue2022PromptCastAN}
Hao Xue and Flora D.Salim.
\newblock Promptcast: A new prompt-based learning paradigm for time series forecasting.
\newblock {\em IEEE Transactions on Knowledge and Data Engineering}, 2022.

\bibitem{Gruver2023LargeLM}
Nate Gruver, Marc Finzi, Shikai Qiu, and Andrew~Gordon Wilson.
\newblock Large language models are zero-shot time series forecasters.
\newblock {\em ArXiv}, abs/2310.07820, 2023.

\bibitem{Li2024UrbanGPTSL}
Zhonghang Li, Lianghao Xia, Jiabin Tang, Yong Xu, Lei Shi, Long Xia, Dawei Yin, and Chao Huang.
\newblock Urbangpt: Spatio-temporal large language models.
\newblock {\em ArXiv}, abs/2403.00813, 2024.

\bibitem{AGCRN}
Lei Bai, Lina Yao, Can Li, Xianzhi Wang, and Can Wang.
\newblock Adaptive graph convolutional recurrent network for traffic forecasting.
\newblock In {\em Proceedings of the 34th International Conference on Neural Information Processing Systems}, NIPS '20, Red Hook, NY, USA, 2020. Curran Associates Inc.

\bibitem{li2018diffusion}
Yaguang Li, Rose Yu, Cyrus Shahabi, and Yan Liu.
\newblock Diffusion convolutional recurrent neural network: Data-driven traffic forecasting.
\newblock In {\em International Conference on Learning Representations}, 2018.

\bibitem{STGCN}
Bing Yu, Haoteng Yin, and Zhanxing Zhu.
\newblock Spatio-temporal graph convolutional networks: a deep learning framework for traffic forecasting.
\newblock In {\em Proceedings of the 27th International Joint Conference on Artificial Intelligence}, IJCAI'18, page 3634–3640. AAAI Press, 2018.

\bibitem{MTGNN}
Zonghan Wu, Shirui Pan, Guodong Long, Jing Jiang, Xiaojun Chang, and Chengqi Zhang.
\newblock Connecting the dots: Multivariate time series forecasting with graph neural networks.
\newblock In {\em Proceedings of the 26th ACM SIGKDD International Conference on Knowledge Discovery \& Data Mining}, KDD '20, page 753–763, New York, NY, USA, 2020. Association for Computing Machinery.

\bibitem{xu2018graph}
Bingbing Xu, Huawei Shen, Qi~Cao, Yunqi Qiu, and Xueqi Cheng.
\newblock Graph wavelet neural network.
\newblock In {\em International Conference on Learning Representations}, 2019.

\bibitem{TGCN}
Ling Zhao, Yujiao Song, Chao Zhang, Yu~Liu, Pu~Wang, Tao Lin, Min Deng, and Haifeng Li.
\newblock T-gcn: A temporal graph convolutional network for traffic prediction.
\newblock {\em IEEE Transactions on Intelligent Transportation Systems}, 21(9):3848--3858, 2020.

\bibitem{Xu2020SpatialTemporalTN}
Mingxing Xu, Wenrui Dai, Chunmiao Liu, Xing Gao, Weiyao Lin, Guo-Jun Qi, and Hongkai Xiong.
\newblock Spatial-temporal transformer networks for traffic flow forecasting.
\newblock {\em ArXiv}, abs/2001.02908, 2020.

\bibitem{Song2020SpatialTemporalSG}
Chao Song, Youfang Lin, S.~Guo, and Huaiyu Wan.
\newblock Spatial-temporal synchronous graph convolutional networks: A new framework for spatial-temporal network data forecasting.
\newblock In {\em AAAI Conference on Artificial Intelligence}, 2020.

\bibitem{Jin2024TimeSF}
Mingyu Jin, Hua Tang, Chong Zhang, Qinkai Yu, Chengzhi Liu, Suiyuan Zhu, Yongfeng Zhang, and Mengnan Du.
\newblock Time series forecasting with llms: Understanding and enhancing model capabilities.
\newblock {\em ArXiv}, abs/2402.10835, 2024.

\bibitem{Yu2023TemporalDM}
Xinli Yu, Zheng Chen, Yuan Ling, Shujing Dong, Zongying Liu, and Yanbin Lu.
\newblock Temporal data meets llm - explainable financial time series forecasting.
\newblock {\em ArXiv}, abs/2306.11025, 2023.

\bibitem{Liu2024LSTPromptLL}
Haoxin Liu, Zhiyuan Zhao, Jindong Wang, Harshavardhan Kamarthi, and B~Aditya Prakash.
\newblock Lstprompt: Large language models as zero-shot time series forecasters by long-short-term prompting.
\newblock {\em ArXiv}, abs/2402.16132, 2024.

\bibitem{Garza2023TimeGPT1}
Azul Garza and Max Mergenthaler-Canseco.
\newblock Timegpt-1.
\newblock {\em ArXiv}, abs/2310.03589, 2023.

\bibitem{Zhou2023OneFA}
Tian Zhou, Peisong Niu, Xue Wang, Liang Sun, and Rong Jin.
\newblock One fits all: Power general time series analysis by pretrained lm.
\newblock In {\em Neural Information Processing Systems}, 2023.

\bibitem{Bian2024MultiPatchPA}
Yuxuan Bian, Xu~Ju, Jiangtong Li, Zhijian Xu, Dawei Cheng, and Qiang Xu.
\newblock Multi-patch prediction: Adapting llms for time series representation learning.
\newblock {\em ArXiv}, abs/2402.04852, 2024.

\bibitem{Liu2024AutoTimesAT}
Yong Liu, Guo Qin, Xiangdong Huang, Jianmin Wang, and Mingsheng Long.
\newblock Autotimes: Autoregressive time series forecasters via large language models.
\newblock {\em ArXiv}, abs/2402.02370, 2024.

\bibitem{Duan2023DeWaveDE}
Yiqun Duan, Jinzhao Zhou, Zhen Wang, Yu~kai Wang, and Ching-Teng Lin.
\newblock Dewave: Discrete eeg waves encoding for brain dynamics to text translation.
\newblock {\em ArXiv}, abs/2309.14030, 2023.

\bibitem{Wimmer2023LeveragingVM}
Christopher Wimmer and Navid Rekabsaz.
\newblock Leveraging vision-language models for granular market change prediction.
\newblock {\em ArXiv}, abs/2301.10166, 2023.

\bibitem{Liu2023ETPLT}
Che Liu, Zhongwei Wan, Sibo Cheng, Mi~Zhang, and Rossella Arcucci.
\newblock Etp: Learning transferable ecg representations via ecg-text pre-training.
\newblock {\em ArXiv}, abs/2309.07145, 2023.

\bibitem{Zhang2024LargeLM}
Xiyuan Zhang, Ranak~Roy Chowdhury, Rajesh~K. Gupta, and Jingbo Shang.
\newblock Large language models for time series: A survey.
\newblock {\em ArXiv}, abs/2402.01801, 2024.

\bibitem{fatemi2024talk}
Bahare Fatemi, Jonathan Halcrow, and Bryan Perozzi.
\newblock Talk like a graph: Encoding graphs for large language models.
\newblock In {\em The Twelfth International Conference on Learning Representations}, 2024.

\bibitem{Zhao2023GraphTextGR}
Jianan Zhao, Le~Zhuo, Yikang Shen, Meng Qu, Kai Liu, Michael Bronstein, Zhaocheng Zhu, and Jian Tang.
\newblock Graphtext: Graph reasoning in text space.
\newblock {\em ArXiv}, abs/2310.01089, 2023.

\bibitem{liu2024one}
Hao Liu, Jiarui Feng, Lecheng Kong, Ningyue Liang, Dacheng Tao, Yixin Chen, and Muhan Zhang.
\newblock One for all: Towards training one graph model for all classification tasks.
\newblock In {\em The Twelfth International Conference on Learning Representations}, 2024.

\bibitem{Chen2023ExploringTP}
Zhikai Chen, Haitao Mao, Hang Li, Wei Jin, Haifang Wen, Xiaochi Wei, Shuaiqiang Wang, Dawei Yin, Wenqi Fan, Hui Liu, and Jiliang Tang.
\newblock Exploring the potential of large language models (llms) in learning on graphs.
\newblock {\em ArXiv}, abs/2307.03393, 2023.

\bibitem{Chen2024LLaGALL}
Runjin Chen, Tong Zhao, Ajay Jaiswal, Neil Shah, and Zhangyang Wang.
\newblock Llaga: Large language and graph assistant.
\newblock {\em ArXiv}, abs/2402.08170, 2024.

\bibitem{Tang2023GraphGPTGI}
Jiabin Tang, Yuhao Yang, Wei Wei, Lei Shi, Lixin Su, Suqi Cheng, Dawei Yin, and Chao Huang.
\newblock Graphgpt: Graph instruction tuning for large language models.
\newblock {\em ArXiv}, abs/2310.13023, 2023.

\bibitem{Liang2023ExploringLL}
Yuebing Liang, Yichao Liu, Xiaohan Wang, and Zhan Zhao.
\newblock Exploring large language models for human mobility prediction under public events.
\newblock {\em ArXiv}, abs/2311.17351, 2023.

\bibitem{Ding2023HiLMDTH}
Xinpeng Ding, Jianhua Han, Hang Xu, Wei Zhang, and X.~Li.
\newblock Hilm-d: Towards high-resolution understanding in multimodal large language models for autonomous driving.
\newblock {\em ArXiv}, abs/2309.05186, 2023.

\bibitem{Dewangan2023Talk2BEVLB}
Vikrant Dewangan, Tushar Choudhary, Shivam Chandhok, Shubham Priyadarshan, Anushka Jain, Arun~Kumar Singh, Siddharth Srivastava, Krishna~Murthy Jatavallabhula, and K.~Madhava Krishna.
\newblock Talk2bev: Language-enhanced bird's-eye view maps for autonomous driving.
\newblock {\em ArXiv}, abs/2310.02251, 2023.

\bibitem{Tanahashi2023EvaluationOL}
Kotaro Tanahashi, Yuichi Inoue, Yu~Yamaguchi, Hidetatsu Yaginuma, Daiki Shiotsuka, Hiroyuki Shimatani, Kohei Iwamasa, Yoshiaki Inoue, Takafumi Yamaguchi, Koki Igari, Tsukasa Horinouchi, Kento Tokuhiro, Yugo Tokuchi, and Shunsuke Aoki.
\newblock Evaluation of large language models for decision making in autonomous driving.
\newblock {\em ArXiv}, abs/2312.06351, 2023.

\bibitem{Zhou2024GPT4VAT}
Xingcheng Zhou and Alois~C. Knoll.
\newblock Gpt-4v as traffic assistant: An in-depth look at vision language model on complex traffic events.
\newblock {\em ArXiv}, abs/2402.02205, 2024.

\bibitem{Jin2023SurrealDriverDG}
Ye~Jin, Xiaoxi Shen, Huiling Peng, Xiaoan Liu, Jin Qin, Jiayang Li, Jintao Xie, Peizhong Gao, Guyue Zhou, and Jiangtao Gong.
\newblock Surrealdriver: Designing generative driver agent simulation framework in urban contexts based on large language model.
\newblock {\em ArXiv}, abs/2309.13193, 2023.

\bibitem{Tan2023LanguageCT}
Shuhan Tan, B.~Ivanovic, Xinshuo Weng, Marco Pavone, and Philipp Kraehenbuehl.
\newblock Language conditioned traffic generation.
\newblock {\em ArXiv}, abs/2307.07947, 2023.

\bibitem{Xu2023UrbanGI}
Fengli Xu, Jun Zhang, Chen Gao, J.~Feng, and Yong Li.
\newblock Urban generative intelligence (ugi): A foundational platform for agents in embodied city environment.
\newblock {\em ArXiv}, abs/2312.11813, 2023.

\bibitem{Zhang2024TransportationGamesBT}
Xue Zhang, Xiangyu Shi, Xinyue Lou, Rui Qi, Yufeng Chen, Jinan Xu, and Wenjuan Han.
\newblock Transportationgames: Benchmarking transportation knowledge of (multimodal) large language models.
\newblock {\em ArXiv}, abs/2401.04471, 2024.

\bibitem{Mumtarin2023LargeLM}
Maroa Mumtarin, Md.~Samiullah Chowdhury, and Jonathan~S. Wood.
\newblock Large language models in analyzing crash narratives - a comparative study of chatgpt, bard and gpt-4.
\newblock {\em ArXiv}, abs/2308.13563, 2023.

\bibitem{Momin2024LeveragingSM}
Khondhaker~Al Momin, Arif~Mohaimin Sadri, and Md~Sami Hasnine.
\newblock Leveraging social media data to identify factors influencing public attitude towards accessibility, socioeconomic disparity and public transportation.
\newblock {\em ArXiv}, abs/2402.01682, 2024.

\bibitem{Wang2023WhereWI}
Xinglei Wang, Meng Fang, Zichao Zeng, and Tao Cheng.
\newblock Where would i go next? large language models as human mobility predictors.
\newblock {\em ArXiv}, abs/2308.15197, 2023.

\bibitem{Mo2023LargeLM}
Baichuan Mo, Hanyong Xu, Dingyi Zhuang, Ruoyun Ma, Xiaotong Guo, and Jinhua Zhao.
\newblock Large language models for travel behavior prediction.
\newblock {\em ArXiv}, abs/2312.00819, 2023.

\bibitem{Zhang2023TrafficGPTVP}
Siyao Zhang, Daocheng Fu, Zhao Zhang, Bin Yu, and Pinlong Cai.
\newblock Trafficgpt: Viewing, processing and interacting with traffic foundation models.
\newblock {\em ArXiv}, abs/2309.06719, 2023.

\bibitem{Da2023OpenTIOT}
Longchao Da, Kuanru Liou, Tiejin Chen, Xuesong Zhou, Xiangyong Luo, Yezhou Yang, and Hua Wei.
\newblock Open-ti: Open traffic intelligence with augmented language model.
\newblock {\em ArXiv}, abs/2401.00211, 2023.

\end{thebibliography}

\end{document}